\documentclass[runningheads]{llncs}
\usepackage[T1]{fontenc}
\usepackage{graphicx}
\usepackage{wrapfig} 
\usepackage{amsmath}
\usepackage{amssymb}		
\usepackage{tabularx}
\usepackage{subfigure}
\usepackage{bbm}
\usepackage{makecell}
\usepackage{hyperref}
\hypersetup{
  colorlinks=true,
  linkcolor=black,
  anchorcolor=black,
  citecolor=black,
  urlcolor=black
}
\usepackage[plainruled,linesnumbered,algochapter]{algorithm2e}

\def\bd{\boldsymbol}        
\DeclareMathOperator*{\argmax}{arg\,max}  
\usepackage{color}

\begin{document}
%
\title{Playing Non-Embedded Card-Based Games with Reinforcement Learning}
%
%
\author{Tianyang Wu \and Lipeng Wan \and Yuhang Wang \and Qiang Wan \and Xuguang Lan\thanks{Corresponding author.}}
\authorrunning{Tianyang Wu et al.}
%
%
\institute{National Key Laboratory of Human-Machine Hybrid Augmented Intelligence,
Institute of Artificial Intelligence and Robotics, Xi'an Jiaotong University, No. 28 West Xianning Road, Xi'an, PR China.}
\maketitle              
\begin{abstract}  

Significant progress has been made in AI for games,
including board games, MOBA, and RTS games.
However, complex agents are typically developed in an embedded manner,
directly accessing game state information, unlike human players who rely on noisy visual data,
leading to unfair competition. Developing complex non-embedded agents remains challenging,
especially in card-based RTS games with complex features and large state spaces.
We propose a non-embedded offline reinforcement learning training strategy
using visual inputs to achieve real-time autonomous gameplay in the RTS game
Clash Royale\footnote{Clash Royale is a trademark of Supercell in Finland and other countries. This content is not approved or sponsored by Supercell.}.
Due to the lack of a object detection dataset for this game, we designed an efficient generative object detection dataset for training.
We extract features using state-of-the-art object detection and optical character recognition models.
Our method enables real-time image acquisition, perception feature fusion, decision-making, and control on mobile devices,
successfully defeating built-in AI opponents. All code is open-sourced at \url{https://github.com/wty-yy/katacr}.
\end{abstract}
\section{Introduction}
Real-Time Strategy (RTS) games typically feature vast state spaces, sparse rewards, incomplete information, and diverse strategies.
Card-based RTS games are a unique type of RTS game where two players use their hands of cards to make real-time decisions.
Unlike typical card games, card-based RTS games are influenced by the complex characteristics of the cards,
as well as the timing and placement of card usage on the battlefield, resulting in an enormous state space.
Players cannot see the cards in the opponent's hand, leading to incomplete information. The outcome of the game is only known at the end of the match,
making rewards very sparse. Strategies such as offense, defense, flexible maneuvering, and tempo control introduce significant uncertainty into the game.

Since Deep Q Network (DQN) \cite{DQN} successfully surpassed human performance on Atari games using deep reinforcement learning (DRL) algorithms,
researchers have attempted to apply DRL to various games, such as board games (AlphaGo \cite{AlphaGo} and AlphaStar \cite{AlphaStar}),
MOBA games (OpenAIFive \cite{OpenAIFive} in DOTA2), and RTS games (AlphaStar \cite{AlphaStar} in StarCraft).
Notably, these agents are embedded models, meaning they can directly obtain accurate state information from the game's underlying mechanics,
which is different from how humans perceive game states through image information. Image data often contains noise that can interfere with decision-making,
leading to competitive inequality due to different state acquisition methods.

Moreover, embedded agents can efficiently interact with the environment, enabling the use of online reinforcement learning algorithms for training,
such as the DQN \cite{DQN} and the Proximal Policy Optimization (PPO) \cite{PPO}. Non-embedded agents,
however, interact with the environment much slower, making online training methods challenging.
Offline reinforcement learning addresses this issue by learning from expert datasets without interacting with the environment,
as demonstrated by Conservative Q-Learning (CQL) \cite{CQL} and Decision Transformer (DT) \cite{DT}.
These algorithms learn policies from expert data and subsequently validate model performance through interactions with the real environment.

\begin{figure}[htbp]
  \centering
  \includegraphics[width=\textwidth]{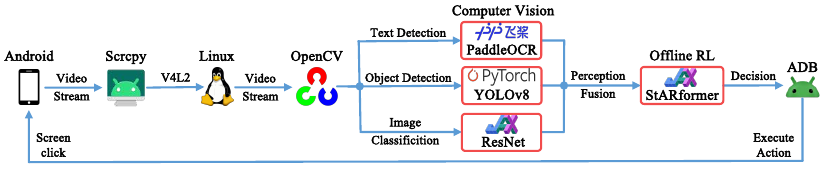}
  \caption{Information Flow Transmission Diagram.}\label{fig-framework}
\end{figure}
\noindent This paper focuses on a popular card-based RTS game, Clash Royale\footnote{\url{https://clashroyale.com/}}(CR).
Specifically, we design a comprehensive interaction process for a non-embedded agent to play CR, which includes interacting with mobile devices,
perceiving and fusing image features, and controlling the agent's decision-making model.
By collecting expert datasets and training the agent using offline reinforcement learning algorithms,
the agent can successfully defeat the top built-in AI in the actual game despite not interacting with the real environment during training.

\section{Card-based Real-Time Strategy Game}
Clash Royale is a popular card-based RTS game, and this paper focuses on the classic one-on-one battle mode.
Both players need to deploy their cards in real-time to defeat their opponent.
The version of the game we tested is from Season 59 in May 2024.

\subsection{Perception Scenarios}
\begin{wrapfigure}[26]{r}{.4\linewidth} 
  \centering\vspace{-5.5ex}
  \includegraphics[width=0.4\textwidth]{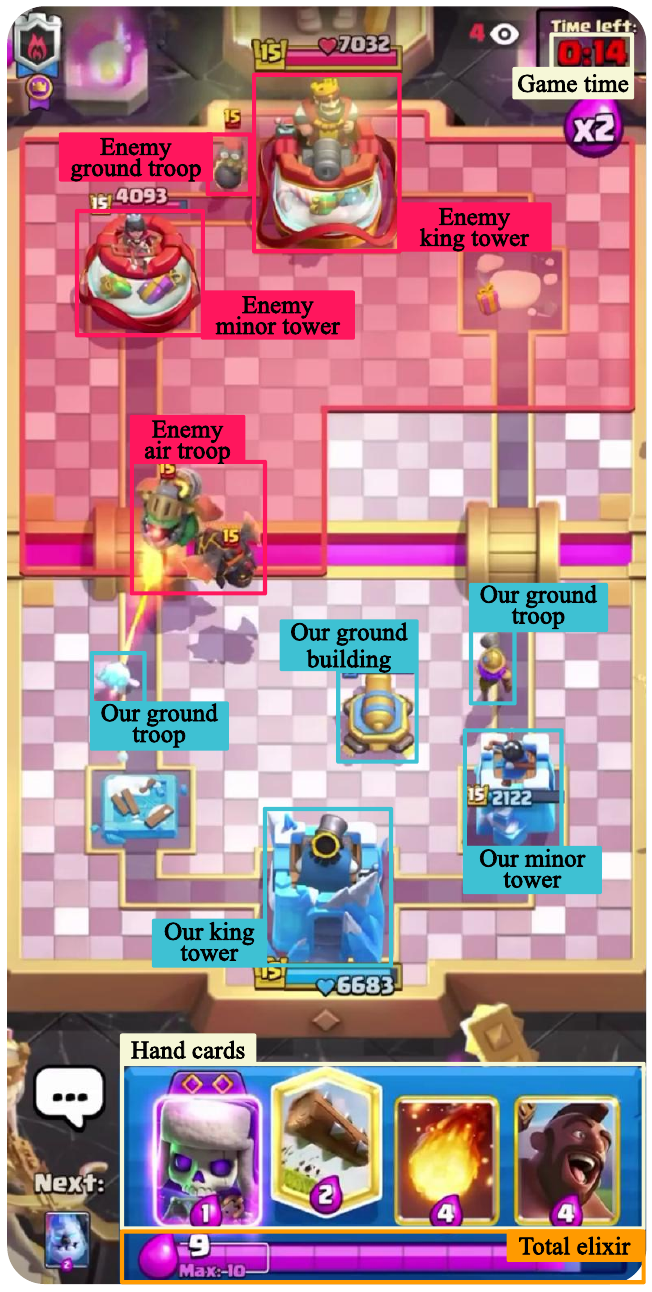}
  \caption{Game Scenario.}\label{fig-introduction}
\end{wrapfigure}
During the match, the game scenario is shown in Fig. \ref{fig-introduction}.
The perceptual content required in this paper is mainly divided into four main parts (arena, hand cards, game time, total elixir)
and three sub-parts within the arena (king tower, minor towers, troops).

The arena includes defensive towers, as well as troops and buildings deployed by both players.
The current game time is displayed at the top right of the arena, indicating the remaining time in the current stage.
At the bottom of the arena are the current hand cards and total elixir.
The hand cards includes the current card images and their elixir costs.
Cards are divided into three types: troops, spells, and buildings, with troops further divided into air and ground types.
Troops are movable units in the arena that can attack and defend;
spells have area effects and can usually quickly kill a group of troops;
buildings are fixed somewhere once used and attack troops in their vicinity.
Total elixir is the resource required to use cards and regenerates automatically over time.

During the game time, players need to destroy as many enemy towers as possible.
In the following subsection, we will specifically introduce the game process
and mathematically model the game states.
\subsection{Game Process}
The game process consists of two parts: card rotation and win conditions.
Card rotation refers to the pattern of card transformation in a player's hand after usage,
while win conditions determine a player's victory, draw, or defeat at the end of a match.

\subsubsection{Card Rotation}
In each match, both players have a deck size of 8 cards. At the beginning of a match,
the cards' appearance order is randomly initialized in a queue. Each time,
4 cards are drawn from the front of the queue. After a player uses a card,
it is returned to the end of the queue. Thus, theoretically, when one side has used all 8 different cards,
the remaining card categories can be logically deduced.
Players need to make real-time decisions on the usage position of card categories based on the current state of the arena,
hand card types, and total elixir information, adopting offensive or defensive strategies.

\subsubsection{Win Conditions}
The target for each player is to destroy as many enemy defensive towers as possible,
with priority given to destroying the opponent's main tower for an immediate victory.
There are different win conditions in two stages of the game:
\begin{enumerate}
\item Regular time: Lasts for three minutes, with the target of destroying more enemy defensive towers.
If the number of defensive towers is equal when time expires, overtime ensues.
Otherwise, the player with more defensive towers wins.
\item Overtime: Lasts for two minutes. In this stage, the player who destroys the remaining enemy defensive towers first wins.
If there is no winner at the end of overtime, the lowest remaining life points of both players' defensive towers are compared.
The player with the higher lowest life points wins. If there is still no winner, the match ends in a draw.
\end{enumerate}

\subsection{Game States}\label{sec-game-state}
In the context of the Arena, the state of defensive towers and troops is characterized by four parameters:
position, health points, class, and faction.

\begin{enumerate}
  \item \textbf{Position}~The position of the $i$-th unit is denoted as a two-dimensional discrete coordinate $\bd{x}_i$,
  representing the unit's location within a grid of size $18 \times 32$ (width $\times$ height).
  \item \textbf{Class}~The class of the $i$-th unit is denoted as a one-dimensional discrete value $c_i$,
  which serves as a unique identifier for each troop or building. The range of possible categories spans from $1$ to $150$.
  \item \textbf{Faction}~The faction of the $i$-th unit is denoted as a binary discrete value $bel_i$,
  indicating the unit's allegiance. The value can be either $0$ or $1$,
  representing friendly and enemy units, respectively.
  \item \textbf{Health}~The health points of the $i$-th unit are represented as a black and white bar image $bar_i$,
  which records the remaining health points of each unit.
\end{enumerate}

For game time, the total elapsed time $t$ can be calculated from the current remaining time.
The duration of a match does not exceed $300$ seconds, hence the range for $t$ is $0 \leq t \leq 300$.

Regarding hand card information, there is a $1$-second delay when a card slot becomes empty after using a card.
When card slots are not empty, the hand card information is composed of the category of cards in each position,
denoted as a four-dimensional discrete coordinate. Since each player can carry up to $8$ cards,
the value in each dimension ranges from $1$ to $8$.

The total elixir information is represented as a one-dimensional discrete value $elixir$.
Given that the maximum elixir limit is $10$, the range for elixir values is $0 \leq elixir \leq 10$.

\section{Generative Dataset}\label{sec-generation-dataset}
For the $i$-th unit $u_i$ in the arena, as defined in Section \ref{sec-game-state},
$u_i$ can be represented as $(\bd{x}_i, c_i, bel_i, bar_i)$.
This section focuses on how to build a generative dataset containing information on $u_i$.

Training object detection models requires a large number of images with bounding box labels,
yet there is currently no object detection dataset available for this game.
Manual frame-by-frame labeling is inefficient and costly. Therefore,
we propose an efficient labeled image generation approach and train it on a restructured YOLOv8 \cite{YOLOv8}.
Subsequently, we validate it on a real video stream dataset,
achieving extremely high accuracy rates as shown in Tab. \ref{table-yolo},
thereby validating the feasibility of the generative dataset.

\begin{figure}[htbp]
  \centering\vspace{-1ex}
  \includegraphics[width=\textwidth]{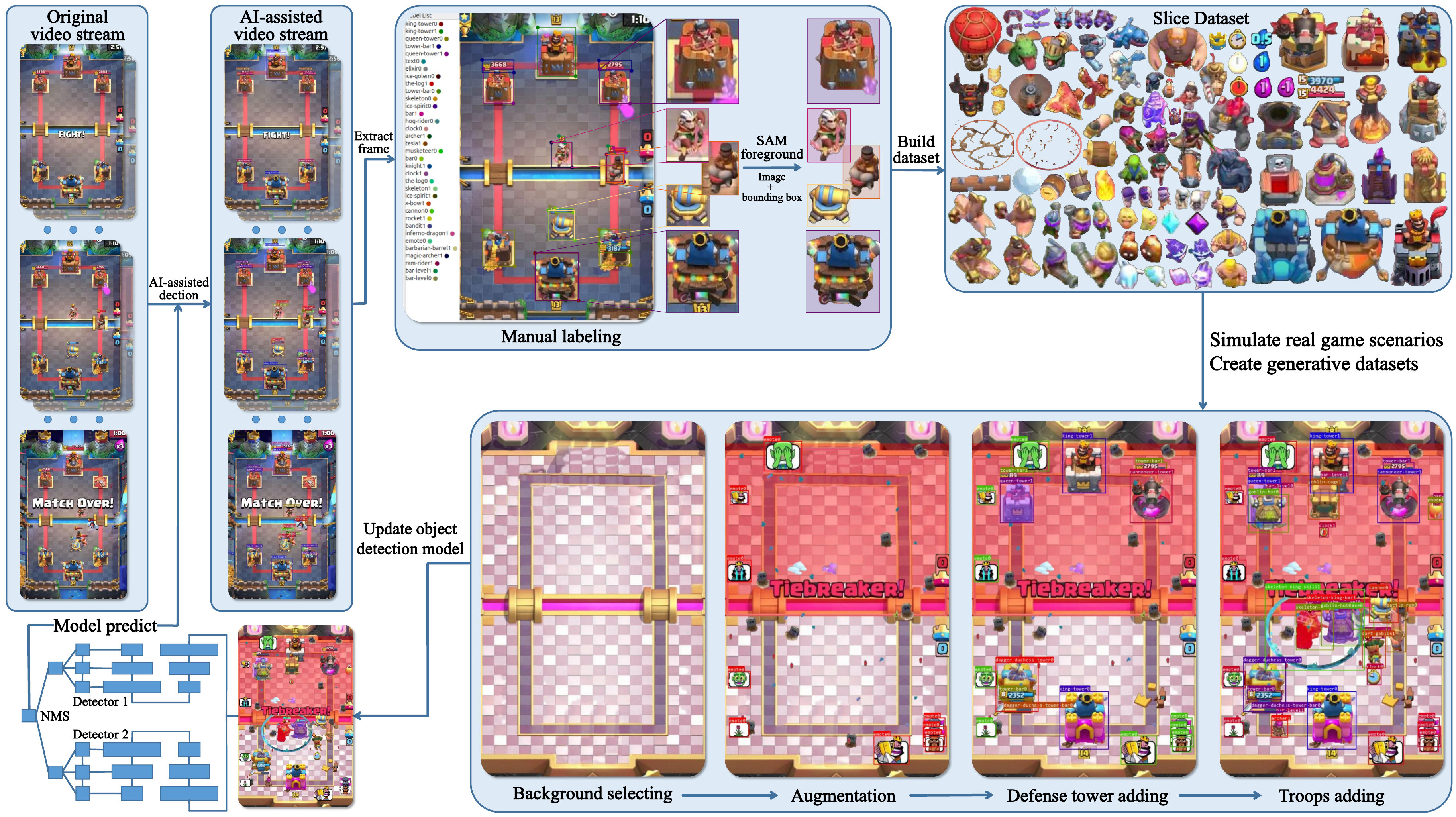}
  \caption{The process of building a generative dataset for object detection.}\label{fig-generation-dataset}
\end{figure}

The generative dataset is based on sliced images of each category unit,  
and the update process of slice image and detection model is illustrated in
Fig. \ref{fig-generation-dataset}. In this process,
the original video stream in the upper-left part represents video data from a episode.
If a previously trained object detection model exists, it is used to assist in labeling the video stream,
resulting in an AI-assisted video stream for further manual labeling.
Otherwise, manual labeling is directly applied to the original video stream.
In manual labeling, a marking interval of 0.5 seconds is used for annotating bounding boxes manually.
Then, both the bounding boxes and the original images are fed into the Segment Anything Model (SAM) \cite{SAM} to obtain foreground segmentation,
and the segmented results are manually filtered to obtain the sliced dataset.
Based on the completed sliced dataset,
the generative dataset algorithm iteratively updates the object detection model for use in the next round of assisted labeling.
\subsection{Generation Algorithm}
Let each slice serve as a drawing unit.
The $i$-th generated object is defined as $g_i = (img_i, box_i, level_i)$,
where $img_i$ corresponds to the slice image of the $i$-th object.
$box_i$ represents the bounding box of the $i$-th object and can be expressed as $(x, y, w, h, c, bel)$,
where $(x, y)$ denotes the coordinates of the center point of the slice in the entire image,
$(w, h)$ represents the width and height of the slice,
$c$ denotes the current slice category, and $bel$ denotes the faction to which the current slice belongs.
$level$ indicates the layer level of the current slice. When generating slices,
they are created sequentially from low to high layers. The layer division is shown in Tab. \ref{tab-level}.

\begin{table}[htbp]
\centering\renewcommand{\arraystretch}{1.2}
\caption{Relationship between Layer Levels and Slice Categories.}
\label{tab-level}
\begin{tabularx}{\textwidth}{>{\centering\arraybackslash\hsize=0.3\hsize}X >{\centering\arraybackslash}X}
\hline
Layer Level & Slice Category \\
\hline
0 & Ground spells, ground background elements \\
1 & Ground troops, defensive towers \\
2 & Air troops, air spells \\
3 & Other detection elements, air background elements \\
\hline
\end{tabularx}
\end{table}

\noindent The insertion process of drawing units is shown in the lower-right part of Fig. \ref{fig-generation-dataset}.
The specific details are as follows:
\begin{enumerate}
\item Background Selection: Randomly select an empty arena image from the dataset, excluding defense towers, troops, and text information, as the background image.
\item Augmentation: Add non-detection object elements in the background board for data augmentation, such as elixir when troops die and randomly appearing butterflies, flowers, etc.
\item Defense Tower Addition: Randomly generate intact or destroyed defense towers at six fixed points on both sides of the arena, and randomly select the associated health information.
\item Troop Addition: Randomly select troop categories inversely proportional to the appearance frequency $\left\{\frac{1}{n_{c_i}-n_{\min}+1}\right\}_{c_i\in C}$,
where $n_{c_i}$ represents the total number of slices generated for category $c_i$,
and $n_{\min}=\min\{n_{c_i}\}_{c_i\in C}$.
Randomly select generation points in the arena according to dynamic probability distribution
and randomly select associated level, health, elixir, and clock information.
\end{enumerate}

\noindent After arranging the drawing units in order, denoted as the sequence $U$,
there might be overlaps among the generated slices. To address this,
we introduce a maximum coverage threshold $\alpha$. If the covered area of a unit exceeds $\alpha$ times the slice area,
the covered unit is removed. After filtering, units are drawn in descending order of layer levels.
The bounding box information of detection categories $C$ is recorded for subsequent model training.
For detailed drawing procedures, refer to Algo. \ref{alg-generator}. 
\vspace{-3ex}
\begin{algorithm}[ht]
	\caption{Pseudocode for Generation Algorithm.\label{alg-generator}}
	\IncMargin{2em}
	\DontPrintSemicolon
	\KwIn{Sequence of drawing units $U=\{u_i\}$, coverage threshold $\alpha$, set of detection categories $C$}
	\KwOut{image, box}
  $\text{image}\gets$ empty image, $\text{box}\gets \{\}$\tcp*{Initialize parameters}
  $U\gets \{u_i\in U: u_i^{level}>u_j^{level}, \forall i, j \in \{1,\cdots,|U|\} \text{~and~} i < j\}$\;
  \While{True}{
    $\text{mask}\gets$ empty mask, $U_{avail}\gets U$\;
    \For{$i=1,2,\cdots,|U|$}{
      \If{$\frac{u^{img}_i\cap~ \text{mask}}{u_i^{img}} > \alpha$}{
        $U_{avail}\gets U_{avail} - R(u_i)$\tcp*{Remove units associated with $u_i$}
      }
      $\text{mask}\gets \text{mask}\cup u_i^{img}$\;
    }
    \If{$|U_{avail}|=|U|$}{
      break\tcp*{Coverage unit filtering completed}
    }
    $U\gets U_{avail}$
  }
  $U\gets \{u_i\in U: u_i^{level}<u_j^{level}, \forall i, j \in \{1,\cdots,|U|\} \text{~and~} i < j\}$\;
  \For{$i=1,2,\cdots,|U|$}{
    $\text{img}\gets \text{img} \cup u_i^{img}$\tcp*{Image drawing}
    \If{$u_{i}^{\text{cls}}\in C$}{
      $\text{box}\gets \text{box} \cup u_i^{box}$\tcp*{Save bounding boxes}
    }
  }
\end{algorithm}
\section{Offline RL Decision Model}
\subsection{Feature Design}
\subsubsection{State}
The model's state input consists of two parts:
$S^{img}$ and $\bd{s}^{card}$. $S^{img}$ is a grid-like feature input of size $\mathbb{R}^{18\times 32\times 15}$,
where each unit's feature $\bd{z}_{ij}$ at the $i$-th row and $j$-th column is a vector in $\mathbb{R}^{15}$.
It encodes four types of information for the unit at that position: category, sub-faction, health, and additional features.
$\bd{s}^{card}$ represents two global features: current hand card information and total holy water amount.
\subsubsection{Action}
The model's action input consists of two parts:
$\bd{a}^{pos}$ and $a^{select}$. $\bd{a}^{pos}$ is a 2D vector representing deployment coordinates,
and $a^{select}$ is the index of the hand card selected for action execution.
\subsubsection{Reward}\label{sec-reward}
The reward design is as follows. Let $h_{i}^{\text{bel}}$ ($i \in \{0,1,2\}, \text{bel} \in \{0,1\}$) be the health points of the defensive towers. When $i=1,2$, it represents the health of the left and right auxiliary towers; $i=0$ represents the health of the main tower. $\text{bel}=0,1$ respectively indicate our and the enemy's buildings. $\Delta h_{i}^{\text{bel}}$ represents the difference in health points between the previous frame and the current frame, and $H_{i}^{\text{bel}}$ represents the total health points of the corresponding defensive towers. The following four reward functions are defined:

1. Defensive Tower Health Reward
\begin{equation}
  r_{tower} = \sum_{\text{bel}=0}^1\sum_{i=0}^2(-1)^{\text{bel}+1}\frac{\Delta h_{i}^{\text{bel}}}{H_{i}^{\text{bel}}}
\end{equation}

2. Defensive Tower Destruction Reward $r_{destory}$: When the enemy's or our auxiliary towers are destroyed, a reward of $(-1)^{\text{bel}+1}$ is given. When the main towers are destroyed, a reward three times that of the auxiliary towers is given.

3. Main Tower Activation Reward $r_{activate}$: When both auxiliary towers are alive, and the main tower loses health points for the first time, a reward of $(-1)^{\text{bel}} \times 0.1$ is given.

4. Elixir Overflow Penalty $r_{elixir}$: When the total elixir continues to overflow, a penalty of 0.05 is given every 1 second.

Combining the above rewards, the total reward is obtained as
\begin{equation}\label{eq-reward}
  r = r_{tower} + r_{destory} + r_{activate} + r_{elixir}
\end{equation}

\subsection{Model Design}
Firstly, we introduce the concepts in reinforcement learning,
considering an infinite horizon discounted Markov Decision Process (MDP),
defined as $(\mathcal{S}, \mathcal{A}, p, r, \gamma)$, where $\mathcal{S}$ is the state space,
$\mathcal{A}$ is the action space,
$p: \mathcal{S} \times \mathcal{A} \times \mathcal{S} \to \mathbb{R}$ is the state transition function,
$r: \mathcal{S} \times \mathcal{A} \to \mathbb{R}$ is the reward function,
and $\gamma \in (0,1)$ is the discount factor.
Let $\pi$ denote the policy function $\pi: \mathcal{S} \times \mathcal{A} \to [0,1]$,
and $R(\pi)$ denote the expected total reward obtained:
\begin{equation}
  R(\pi) = \mathbb{E}_{S_1, A_1, S_2, A_2 \cdots}\left[\sum_{t=0}^{\infty} r(S_t, A_t)\right],\quad
  \text{where } A_t \sim \pi(\cdot|S_t), S_{t+1} \sim p(\cdot|S_t, A_t)
\end{equation}
The goal of reinforcement learning is typically to find the optimal policy $\pi^* := \argmax_{\pi} R(\pi)$.
Online reinforcement learning algorithms often update policies through policy iteration and value function estimation.
In this paper, we use an offline reinforcement learning algorithm that is more akin to imitation learning rather than value estimation methods.

Decision Transformer (DT) \cite{DT} is a method that formulates the reinforcement learning problem
as a sequence modeling problem using the Transformer \cite{transformer} architecture from deep learning.
For an interaction trajectory $\rho$ of length $T$ in the offline dataset:
\begin{equation}
  \rho = (s_1, a_1, r_1, s_2, a_2, r_2, \cdots, s_T, a_T, r_T, s_{T+1})
\end{equation}
where $s_{T+1}$ is the terminal state, $\rho$ can be modeled as a sequence:
\begin{equation}\label{eq-sequence}
  R_0, s_1, a_1, R_1, s_2, \cdots, a_{T-1}, R_{T-1}, s_T, a_T
\end{equation}
where $R_i = \sum_{t=i}^T r_{t+1}, (i=0, \cdots, T-1)$ represents the return-to-go.

\noindent The sequence encoding model in DT uses the GPT-1 \cite{GPT},
which only contains the encoder part of the causal attention mechanism.
Causal attention (each feature $i$ can only see features $j \leq i$) is defined as:
\begin{equation}\label{eq-attn}
\begin{aligned}
  Z = \text{softmax}\left(\frac{\left(QK^T\right) \odot M}{\sqrt{d_k}}\right)V
\end{aligned}
\end{equation}
where $M$ is the mask matrix, an $N$-order lower triangular matrix.
$Q$, $K$, and $V$ represent the query, key, and value generated from the input sequence $X$, respectively.
The query and key have the same feature dimension $d_k$.
When $M$ in Eq. \eqref{eq-attn} is an all-ones matrix, $Z$ is defined as cross attention.

\begin{figure}[htbp]
  \centering
  \includegraphics[width=\textwidth]{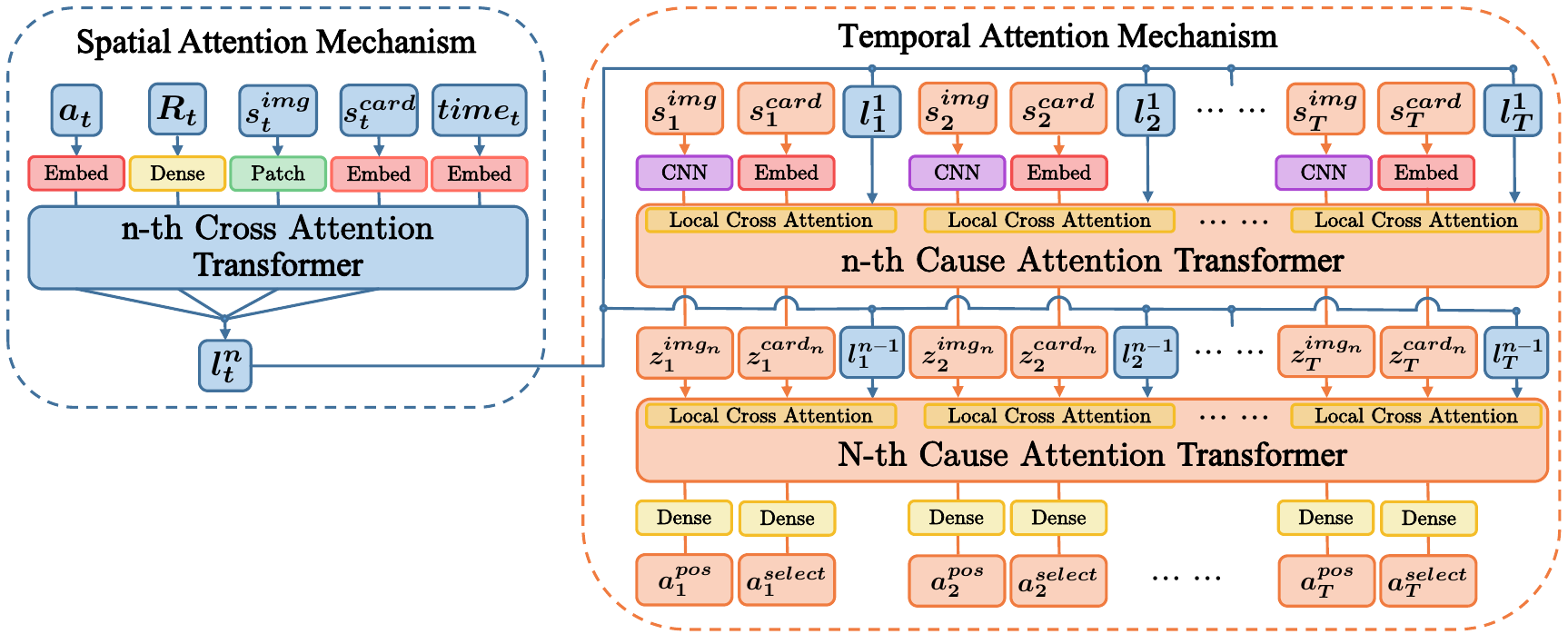}
  \caption{Decision model architecture: A spatial attention mechanism on the left encodes feature information
  at the same timestep, while a temporal attention mechanism on the right associates information across consecutive
  frames to predict actions.
  }\label{fig-policy-model}
\end{figure}

\noindent The model we designed is shown in Fig. \ref{fig-policy-model},
based on the architecture of StARformer \cite{StARformer}.
The model input is the trajectory sequence $(a_t, R_t, s_t)_{t=1}^T$,
and the output is the action prediction sequence $(a_t^{pos}, a_t^{select})_{t=1}^T$.
The left side uses cross attention to encode local information $(a_t, R_t, s_t)$ by spatial dimension,
and the image $s_t^{img}$ is converted into a sequence using the patch method in Vision Transformer (ViT) \cite{ViT}.
The right side uses causal attention to encode global information $(s^{img}_t, s^{card}_t)$ by temporal dimension,
and corresponding local encoding information $l_t^n$ is introduced in each layer's sequence input.

To enable mutual attention among information at the same moment,
local cross-attention is introduced. Specifically,
the implementation involves defining the masking matrix \( M \) in Eq. \eqref{eq-attn} as \( M_{3} \).
Here, $M_{L_0}$ is defined as follows, consider a trajectory input of length $L$, then
\begin{equation}
  (M_{L_0})_{ij} = \begin{cases}
    1, & i=kL_0-l, j \leq kL_0 \\
    0, & \text{otherwise}
  \end{cases},\quad k \in \{1, \cdots, L\}, l \in \{0, \cdots, L_0-1\}
\end{equation}
Let $\mathbbm{H}$ represents a placeholder. We designed three model architectures:
\begin{enumerate}
  \item \textbf{StARformer-3L}: The input sequence length is $3L$.
  The temporal sequence output of the $n$-th layer Cause Attention is $\{z_t^{img_n}, z_t^{card_n}, l_t^{n-1}\}_{t=1}^L$,
  with a local attention mask matrix $M_3$. The output is $(\hat{a}_t^{pos}, \hat{a}_t^{select}, \mathbbm{H})_{t=1}^T$.
  \item \textbf{StARformer-2L}: The model input sequence length is $2L$.
  The temporal sequence output of the $n$-th layer Cause Attention is $\{z_t, l_t^{n-1}\}_{t=1}^L$,
  with a local attention mask matrix $M_2$. The output corresponds to $(\left[\hat{a}_t^{pos}, \hat{a}_t^{select}\right], \mathbbm{H})_{t=1}^T$.
  \item \textbf{DT-4L}: The model input length is $4L$, containing only the Cause Attention in the temporal attention mechanism.
  The temporal sequence output of the $n$-th layer is $\{a_{t-1}, R_{t-1}, s_t^{img_n}, s_t^{card_n}\}_{t=1}^L$.
  The output corresponds to $(\mathbbm{H}, \mathbbm{H}, \hat{a}_t^{pos}, \hat{a}_t^{select})_{t=1}^T$.
\end{enumerate}

\subsection{Prediction Target Design and Resampling}
Due to the highly discrete nature of action execution in this task, only 4\% of the total frames are action execution frames,
while the remaining frames do not execute any actions. Directly predicting actions frame by frame would result in a severe long-tail problem,
causing the model to rarely execute actions (as seen in Tab. \ref{table-model-eval},
where the number of actions predicted in a discrete manner is far lower than those predicted continuously).
Therefore, it is necessary to transform the prediction target from discrete to continuous.
The solution is to introduce delayed action prediction: for the $i$-th frame,
find the nearest subsequent action frame $j$ (including itself), and let the maximum interval frame threshold be $T_{delay}$.
The delayed action prediction for each non-action frame is then $a^{delay}_{i} = \min\{j-i, T_{delay}\}$.

\begin{figure}[htbp]
\centering
\includegraphics[width=\textwidth]{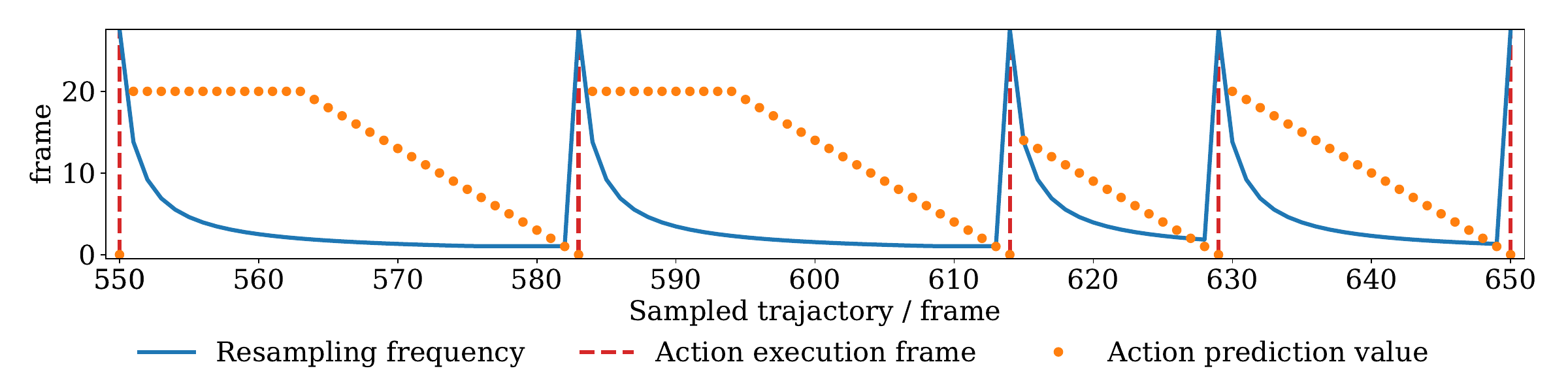}
\caption{A segment of data extracted from the offline dataset,
containing a total of 5 action frames, with a maximum interval frame threshold $T_{delay} = 20$.}\label{fig-resample-and-delay}
\end{figure}
\noindent When sampling the offline dataset, to avoid model bias caused by the long-tail problem,
this paper also sets the resampling frequency.
Let the total number of frames in the dataset be $N$ and the number of action frames be $N_{action}$,
then the action frame ratio is $r_a := N_{action} / N$.
For the $i$-th action frame located at the $t_i$-th frame in the dataset,
the resampling frequency for the ending frames of the trajectory $j \in \{t_{i}, \cdots, t_{i+1}-1\}$ is given by
\begin{equation}\label{eq-resample-freq}
s_j = \max\left\{\frac{1}{1-r_a}, \frac{1}{r_a(j-t_i+1)}\right\}, \quad (t_i \leqslant j \leqslant t_{i+1})
\end{equation}
Thus, the sampling distribution of the ending frames in the training trajectory is $\left\{s_j/\left(\sum_{j=1}^{N}s_j\right)\right\}_{j=1}^N$.
Fig. \ref{fig-resample-and-delay} shows the resampling frequency and action prediction values corresponding to a segment of the trajectory from the offline dataset.

\section{Data Analysis and Experimental Results}

\subsection{Generative Dataset Analysis}\label{sec-exp-data}
\begin{figure}[htbp]
  \centering
  \includegraphics[width=\textwidth]{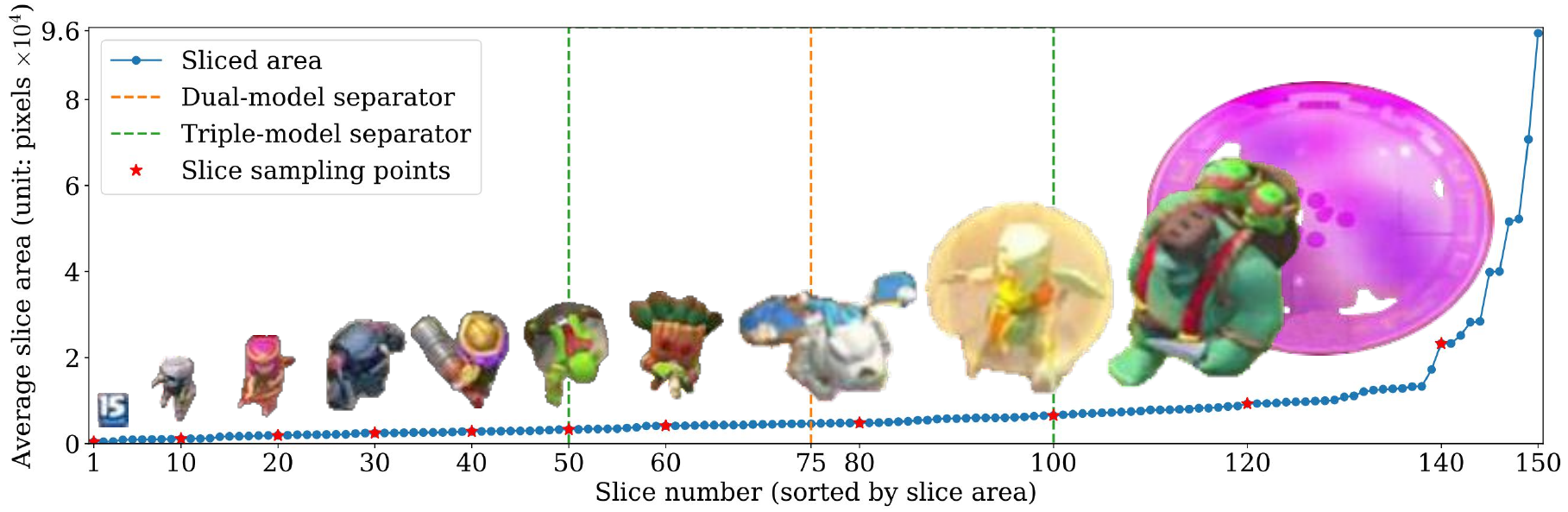}
  \caption{All dataset slices are sorted and numbered by their average areas.
  The bisect and trisect separators for the dual and triple models, respectively.}
  \label{fig-segment}
\end{figure}

The dataset is divided into two parts\footnote{The dataset statistics are accurate as of May 6, 2024, and all image datasets have been open-sourced: \url{https://github.com/wty-yy/Clash-Royale-Detection-Dataset}}:
\begin{enumerate}
  \item Generative dataset slices: a total of 154 categories, with 150 categories to be detected, totaling 4654 slices.
  The distribution of sizes among all slices of the categories to be detected is shown in Fig. \ref{fig-segment}.
  \item Target detection validation set: a total of 6939 manually labeled target detection images,
  containing 116878 target boxes. On average, each image contains 17 target boxes.
  This dataset is obtained by frame-by-frame labeling of real game video streams,
  while the models are trained solely on the generative dataset, so this dataset can be used as a validation set.
\end{enumerate}

\subsection{Object Detection Models}\label{sec-exp-detect}
The object detection models used the self-implemented YOLOv5 and the refactored YOLOv8 models.
Each training dataset size was set to 20,000, and at most 80 epochs were trained for convergence.
Data augmentation included HSV augmentation, image rotation, random horizontal and vertical translation,
image scaling, and horizontal flipping.

The experimental results are shown in Tab. \ref{table-yolo}, where the letters after the model name represent the model size,
with "l" and "x" corresponding to large and extra-large models, respectively.
YOLOv8-l$\times n$ indicates the use of $n$ YOLOv8-l models,
with each sub-model detecting slice types within the regions delimited by the separators in Fig. \ref{fig-segment}.

FPS indicates the model's validation speed tested on GeForce RTX 4090 with 1 batch size.
FPS(T) represents the detection speed with ByteTrack target tracking \cite{ByteTrack} on GeForce RTX 4060 Laptop.


The experimental results show that the YOLOv8-l$\times 2$ performs almost as well as the $\times 3$ for small targets and significantly better than non-combined detectors.
This may be because the 150 predicted categories exceed the model's detection capacity,
and there's a significant size difference between the largest and smallest bounding boxes.
Since YOLOv8 uses anchor-free detection heads, it detects large targets more easily,
leading to generally larger predicted bounding boxes.
Therefore, using multiple detectors to reduce the average number of categories can improve small target detection.

\begin{table}[!h]
	\renewcommand{\arraystretch}{1.2}
	\centering
	\caption{Comparison of YOLO Models.} \label{table-yolo}
	\begin{tabularx}{\textwidth} { 
   >{\centering\arraybackslash}X 
   >{\centering\arraybackslash\hsize=.3\hsize}X
   >{\centering\arraybackslash\hsize=.3\hsize}X
   >{\centering\arraybackslash\hsize=.3\hsize}X
   >{\centering\arraybackslash\hsize=.3\hsize}X
   >{\centering\arraybackslash\hsize=.5\hsize}X
   >{\centering\arraybackslash\hsize=.3\hsize}X
   >{\centering\arraybackslash\hsize=.5\hsize}X
   >{\centering\arraybackslash\hsize=.9\hsize}X
  }
  \hline
  Model Name&AP50&P50&R50&mAP&mAP(S)&FPS&FPS(T)&Augmentation\\
  \hline
  YOLOv5-l&66.2&84.4&63.8&53.2&NA&59&NA&\\
  YOLOv8-x&83.1&\textbf{93.9}&68.3&67.7&39.8&68&31&\\
  YOLOv8-x&\textbf{85.3}&90.7&80.4&66.8&35.9&68&31&\checkmark\\
  YOLOv8-l~$\times 2$&84.3&89.5&79.8&67.4&43.9&34&18&\checkmark\\
  YOLOv8-l~$\times 3$&85.2&89.7&\textbf{80.9}&\textbf{68.8}&\textbf{48.3}&23&10&\checkmark\\
  \hline
	\end{tabularx}
\end{table}

\subsection{Decision Model}\label{sec-decision-model}
We manually constructed expert data\footnote{Expert dataset: \url{https://github.com/wty-yy/Clash-Royale-Replay-Dataset}}
consisting of 105 episodes between a player and a built-in AI with 8000 rating,
with both sides using fixed card decks.
The dataset consists of a total of 113,981 frames, with action frames accounting for 4.01\%,
and the resampling frequency ratio being $\frac{\text{action frames}}{\text{non-action frames}} = 24.92:1.04$.
The average action delay is 21.26 frames, and the maximum interval frame threshold is $T_{delay}=20$.
\begin{enumerate}
  \item Resampling: Resampling sparse action frames proportionally to accelerate model convergence and alleviate the long-tail problem of offline datasets.
  \item Random card reshuffling: Shuffling all cards in the current input trajectory according to a random permutation. When predicting all card indices, the cards corresponding to actions are also transformed accordingly.
\end{enumerate}

\noindent Tab. \ref{table-model-eval} records the highest average reward obtained from 20 rounds of interaction with the environment for the first 10 epochs of each model.
Each column in table are as follow, total reward represents the cumulative reward obtained according to the reward function Eq. \eqref{eq-reward}.
Game duration refers to the duration of each game episode.
Number of actions represents the total number of actions successfully executed by the agent in each episode.
We observed a 37\% improvement in performance when transitioning from discrete prediction to continuous prediction.
Additionally, enhancing the StARformer architecture from 2L to 3L resulted in a 24\% increase in model performance.
\begin{table}[!h]
	\renewcommand{\arraystretch}{1.2}
	\centering
	\caption{Comparison of Decision Models.}\label{table-model-eval}
	\begin{tabularx}{\textwidth} { 
    >{\centering\arraybackslash\hsize=1.6\hsize}X 
    >{\centering\arraybackslash\hsize=0.6\hsize}X 
    >{\centering\arraybackslash\hsize=0.9\hsize}X
    >{\centering\arraybackslash}X
    >{\centering\arraybackslash}X 
    >{\centering\arraybackslash\hsize=.5\hsize}X
    }
  \hline
	Model Architecture&Step Size $L$&Total Reward&Game Duration&Number of Actions&Win Rate\\
  \hline
    DT-4L&$50$&$-5.7\pm 2.5$&$148.9\pm 33.6$&$128.7\pm 37.7$&$5\%$\\
    StARformer-2L&$30$&$-6.0\pm 2.3$&$135.0\pm 35.1$&$141.8\pm 57.9$&$5\%$\\
    StARformer-2L&$50$&$-6.2\pm 2.2$&$131.9\pm 44.3$&$195.3\pm 69.8$&$5\%$\\
    StARformer-2L&$100$&$-4.9\pm 2.8$&$150.2\pm 35.6$&$187.6\pm 48.2$&$0\%$\\
    StARformer-3L&$30$&$-5.1\pm 3.7$&$147.2\pm 37.4$&$190.8\pm 52.7$&$\mathbf{10\%}$\\
    StARformer-3L&$50$&$\mathbf{-4.7\pm 3.1}$&$\mathbf{158.9\pm 27.7}$&$\mathbf{207.8\pm 48.2}$&$5\%$\\
    StARformer-3L&$100$&$-6.1\pm 2.2$&$125.9\pm 37.8$&$144.6\pm 42.9$&$5\%$\\
    \makecell[c]{StARformer-3L\\(Full Card)}&$50$&$-5.6\pm 2.1$&$150.2\pm 38.6$&$195.3\pm 69.8$&$0\%$\\
    \makecell[c]{StARformer-2L\\(Discrete Action)}&$50$&$-7.5\pm 0.8$&$123.1\pm 39.2$&$21.9\pm 9.4$&$0\%$\\
  \hline
	\end{tabularx}
\end{table}

\noindent The validation environment includes a smartphone with HarmonyOS,
a computer system running Ubuntu 24.04 LTS, with an R9 7940H CPU and an RTX GeForce 4060 Laptop GPU.
The average decision-making time is 120ms, and the perception fusion time is 240ms.


\section{Conclusion}

In this paper, we propose a novel non-embedded offline reinforcement learning training strategy based on the game Clash Royale.
By integrating state-of-the-art algorithms for object detection and optical character recognition,
we successfully enable agents to play real-time matches on mobile devices and defeat the built-in AI opponents.

We provide new insights into the application of offline reinforcement learning on mobile devices.
Future work can focus on algorithmic improvements. Currently,
training is conducted with a fixed set of cards and our agents still cannot consistently defeat the built-in AI,
falling short of human-level performance. Moreover,
creating offline reinforcement learning datasets requires substantial human effort.
To further enhance the capabilities of our agents, online reinforcement learning algorithms should be considered.
Additionally, more efficient perception fusion algorithms and decision model architectures are needed to improve real-time decision-making and match win rates.

All code used in this work has been open-sourced\footnote{All code: \url{https://github.com/wty-yy/katacr}}.
Videos of the agent's victories have been uploaded\footnote{Match videos: \url{https://www.bilibili.com/video/BV1xn4y1R7GQ}}.
We hope this work provides valuable reference and inspiration for researchers in related fields.

\begin{credits}
\subsubsection{\ackname} This work was supported in part by NSFC under grant No.62125305, No. U23A20339, No.62088102,
No. 62203348.
\end{credits}

\bibliographystyle{splncs04}
\bibliography{mybib}
\end{document}